# Generalized Neutrosophic Soft Set

Said Broumi

Faculty of Arts and Humanities, Hay El Baraka Ben M'sik Casablanca B.P. 7951, Hassan II Mohammedia-Casablanca University, Morocco

broumisaid78@gmail.com

## Abstract

*In this paper we present a new concept called "generalized neutrosophic soft set". This concept incorporates the beneficial properties of both generalized neutrosophic set introduced by A.A.Salama [7] and soft set techniques proposed by Molodtsov [4]. We also study some properties of this concept. Some definitions and operations have been introduced on generalized neutrosophic soft set. Finally we present an application of generalized neuutrosophic soft set in decision making problem.*

## 1. INTRODUCTION

In Many complicated problems like, engineering problems, social, economic, computer science, medical science…etc, the data associated are not necessarily crisp, precise, and deterministic because of their vague nature. Most of these problem were solved by different theories. One of these theories was the fuzzy set theory discovered by Lotfi, Zadeh in 1965 [1], Later several researches present a number of results using different direction of fuzzy set such as: interval fuzzy set [12], generalized fuzzy set by Atanassov [2]..., all these are successful to some extent in dealing with the problems arising due to the vagueness present in the real world ,but there are also cases where these theories failed to give satisfactory results, possibly due to indeterminate and inconsistent information which exist in belief system, then in 1995, Smarandache [3] initiated the theory of neutrosophic set as new mathematical tool for handling problems involving imprecise, indeterminacy, and inconsistent data. Several researchers dealing with the concept of neutrosophic set such as M. Bhowmik and M.Pal in [13], A.A.Salama in [7], and H.Wang in [14]. Furthermore, In 1999, a Russian mathematician (Molodtsov [4]) introduce a new mathematical tool for dealing with uncertainties, called "soft set theory". This new concept is free from the limitation of variety of theories such as probability theory, Fuzzy sets and rough sets. Soft set theory has no problem of setting the membership function, which makes it very convenient and easy to apply in practice. After Molodtsov'work, there have been many researches in combining fuzzy set with soft set, which incorporates the beneficial properties of both fuzzy set and soft set techniques ( see [11] [6] [8]). So in this paper we present a new model which combine two concepts: Generalized neutrosophic set proposed by A.A.Salama [7] and soft set proposed by Molodtsov in [4], together by introducing a new concept called generalized neutrosophic sof set, thus we introduce its operations namely equal, subset, union, and intersection, Finally we present an application of generalized neutrosophic soft set in decision making problem.

The content of the paper is orgonaized as follow: In section 2, we briefly present some basic definitions and preliminary results are given which will be used in the rest of the paper. In section 3, generalized neutrosophic soft set. In section 4 an application of generalized neutrosophic soft set in a decision making problem. Finally section 5 presents the conclusion of our work.

## 2 .Preliminaries

In this section, we review some definitions with regard to neutrosophic set, generalized neutrosophic set and and soft set. The definitions in this part may be found in references [3, 4, 7, 10].

Throughout this paper, let U be a universal set and E be the set of all possible parameters under consideration with respect to U, usually, parameters are attributes, characteristics, or properties of objects in U.

### Definition 2.1 (see [3]). Neutrosophic Set

Let U be an universe of discourse then the neutrosophic set A is an object having the form A = {< x: $T_{A(x)}, I_{A(x)}, F_{A(x)}$ >,x $\in$ U}, where the functions T,I,F : U→]⁻0,1⁺[ define respectively the degree of membership , the degree of indeterminacy, and the degree of non-membership of the element x $\in$ X to the set A with the condition:

$$^-0 \leqslant T_{A(x)} + I_{A(x)} + F_{A(x)} \leqslant 3^+.$$

From philosophical point of view, the neutrosophic set takes the value from real standard or non-standard subsets of ]⁻0,1⁺[. So instead of ]⁻0,1⁺[ we need to take the interval [0,1] for technical applications, because ]⁻0,1⁺[ will be difficult to apply in the real applications such as in scientific and engineering problems.

### Definition 2.2. (see [10])

A neutrosophic set A is contained in another neutrosophic set B i.e. A ⊆ B if x ∈ U, $T_A(x)$ ≤ $T_B(x)$, $I_A(x)$ ≤ $I_B(x)$, $F_A(x)$ ≥ $F_B(x)$.

### Definition 2.3(see [7]). Generalized Neutrosophic Set

Let X be a non-empty fixed set. A generalized neutrosophic set (GNS for short) A is an object having the form A = {< x: $T_{A(x)}, I_{A(x)}, F_{A(x)}$ >,x ∈ U}, Where $T_{A(x)}$, represent the degree of membership function , $I_{A(x)}$ represent the degree of indeterminacy, and $F_{A(x)}$ represent the degree of non-member ship respectively of each element x ∈ X to the set A where the functions satisfy the condition:

$$T_{A(x)} \land F_{A(x)} \land I_{A(x)} \leq 0.5$$

As an illustration, let us consider the following example.

**Example2.4.** Assume that the universe of discourse U={$x_1,x_2,x_3$},where $x_1$ characterizes the capability, $x_2$ characterizes the trustworthiness and $x_3$ indicates the prices of the objects. It may be further assumed that the values of $x_1$, $x_2$ and $x_3$ are in [0, 1] and they are obtained from some questionnaires of some experts. The experts may impose their opinion in three components viz.

the degree of goodness, the degree of indeterminacy and that of poorness to explain the characteristicsof the objects. Suppose A is an generalized neutrosophic set ( GNS ) of U, such that, A = {< $x_1$, 0.3, 0.5, 0.4 >,< $x_2$,0.4, 0.2, 0.6 >,< $x_3$, 0.7, 0.3, 0.5 >}, where the degree of goodness of capability is 0.3, degree of indeterminacy of capability is 0.5 and degree of falsity of capability is 0.4 etc.

**Definition 2.5** (see[4]). **Soft Set**

Let U be an initial universe set and E be a set of parameters. Let P(U) denotes the power set of U. Consider a nonempty set A, A ⊂ E A pair ( F, A ) is called a soft set over U, where F is a mapping given by F : A → P(U).

As an illustration , let us consider the following example.

**Example 2.6.** Suppose that U is the set of houses under consideration, say U = {$h_1$, $h_2$, . . ., $h_5$}. Let E be the set of some attributes of such houses, say E = {$e_1$, $e_2$, . . ., $e_8$}, where $e_1$, $e_2$, . . ., $e_8$ stand for the attributes "expensive", "beautiful", "wooden", "cheap", "modern", and "in bad repair", respectively.

In this case, to define a soft set means to point out expensive houses, beautiful houses, and so on. For example, the soft set (F, A) that describes the "attractiveness of the houses" in the opinion of a buyer, say Thomas, may be defined like this:

A={$e_1$, $e_2$, $e_3$, $e_4$, $e_5$};
F($e_1$) = {$h_2$, $h_3$, $h_5$}, F($e_2$) = {$h_2$, $h_4$}, F($e_3$) = {$h_1$}, F($e_4$) = U, F($e_5$) = {$h_3$, $h_5$}.

For more details on the algebra and operations on generalized neutrosophic set and soft set, the reader may refer to [ 5, 6, 8 , 7, 9, 11].

# 3. Generalized Neutrosophic Soft Set

In this section ,we will initiate the study on hybrid structure involving both generalized neutrosophic set and soft set theory.

### Definition 3.1

Let U be an initial universe set and A    E be a set of parameters. Let GNS( U ) denotes the set of all generalized neutrosophic sets of U. The collection (F,A) is termed to be the soft generalized neutrosophic set over U, where F is a mapping given by F : A      GNS(U).

**Remark 3.2**. We will denote the generalized neutrosophic soft set defined over an universe by GNSS.

Let us consider the following example.

### Example 3.3

 Let U be the set of blouses under consideration and E is the set of parameters (or qualities). Each parameter is a generalized neutrosophic word or sentence involving generalized neutrosophic words. Consider E = { Bright, Cheap, Costly, very costly, Colorful, Cotton, Polystyrene, long sleeve , expensive }. In this case, to define a generalized neutrosophic soft set means to point out

Bright blouses, Cheap blouses, Blouses in Cotton and so on. Suppose that, there are five blouses in the universe U given by, U = {$b_1$, $b_2$, $b_3$, $b_4$, $b_5$} and the set of parameters A = {$e_1$, $e_2$, $e_3$, $e_4$}, where each $e_i$ is a specific criterion for blouses:

$e_1$ stands for 'Bright',
$e_2$ stands for 'Cheap',
$e_3$ stands for 'costly',
$e_4$ stands for 'Colorful',

Suppose that,

F(Bright)={<$b_1$, 0.5, 0.6, 0.3>,<$b_2$,0.4, 0.7, 0.2>,<$b_3$,0.6, 0.2, 0.3>,<$b_4$, 0.7, 0.3, 0.2> ,<$b_5$,0.8, 0.2, 0.3>}.

F(Cheap)={<$b_1$, 0.6, 0.3, 0.5>,<$b_2$, 0.7, 0.4, 0.3>,<$b_3$, 0.8, 0.1, 0.2>,<$b_4$, 0.7, 0.1, 0.3> ,<$b_5$, 0.8, 0.3, 0.4}.

F(Costly)={<$b_1$, 0.7, 0.4, 0.3>,<$b_2$, 0.6, 0.1, 0.2>,<$b_3$, 0.7, 0.2, 0.5>,< $b_4$, 0.5, 0.2, 0.6 > ,< $b_5$, 0.7, 0.3, 0.2 >}.

F(Colorful)={<$b_1$, 0.8, 0.1, 0.4>,<$b_2$, 0.4, 0.2, 0.6>,<$b_3$, 0.3 , 0.6, 0.4>,<$b_4$, 0.4,0.8,0.5> ,< $b_5$, 0.3, 0.5, 0.7 >}.

The generalized neutrosophic soft set ( GNSS ) ( F, E ) is a parameterized family {F($e_i$), i = 1,···,10} of all generalized neutrosophic sets of U and describes a collection of approximation of an object. The mapping F here is 'blouses (.)', where dot(.) is to be filled up by a parameter $e_i$ ∈ E. Therefore, F($e_1$) means 'blouses (Bright)' whose functional-value is the generalized neutrosophic set {< $b_1$, 0.5, 0.6, 0.3 >,< $b_2$, 0.4, 0.7, 0.2 >, < $b_3$, 0.6, 0.2, 0.3 >,< $b_4$, 0.7, 0.3, 0.2 >,< $b_5$, 0.8, 0.2, 0.3 >}.

Thus we can view the generalized neutrosophic soft set ( GNSS ) (F,A) as a collection of approximation as below:

( F, A ) = { Bright blouses= {< $b_1$, 0.5, 0.6, 0.3 >,< $b_2$, 0.4, 0.7, 0.2 >, < $b_3$, 0.6, 0.2, 0.3 >,< $b_4$,0.7,0.3,0.2 >,< $b_5$,0.8,0.2,0.3 >}, Cheap blouses= {< $b_1$,0.6,0.3,0.5 >,< $b_2$,0.7,0.4,0.3 >,< $b_3$,0.8,0.1,0.2 >, < $b_4$, 0.7,0.1,0.3 >,< $b_5$, 0.8,0.3,0.4 >}, costly blouses= {< $b_1$,0.7,0.4,0.3 > ,< $b_2$, 0.6, 0.1, 0.2 >,< $b_3$, 0.7, 0.2, 0.5 >,< $b_4$, 0.5,0.2,0.6 >,< $b_5$, 0.7, 0.3, 0.2 >}, Colorful blouses= {< $b_1$, 0.8, 0.1,0.4 >,< $b_2$, 0.4, 0.2,0.6 >,< $b_3$, 0.3, 0.6,0.4 >, < $b_4$, 0.4, 0.8, 0.5>,< $b_5$, 0.3, 0.5, 0.7 >}}. where each approximation has two parts: (i) a predicate p, and (ii) an approximate value-set v ( or simply to be called value-set v ).

For example, for the approximation 'Bright blouses={<$b_1$,0.5,0.6,0.3>,<$b_2$,0.4,0.7,0.2 >,<$b_3$,0.6,0.2,0.3>,<$b_4$,0.7,0.3,0.2>,<$b_5$,0.8,0.2,0.3>}'.

We have (i) the predicate name 'Bright blouses', and (ii) the approximate value-set is{<$b_1$,0.5,0.6,0.3>,<$b_2$,0.4,0.7,0.2>,<$b_3$,0.6,0.2,0.3>,<$b_4$,0.7,0.3,0.2> ,< $b_5$,0.8,0.2,0.3 >}. Thus, an generalized neutrosophic soft set ( F, E ) can be viewed as a collection of approximation like ( F, E ) = {$p_1$ = $v_1$,$p_2$ = $v_2$,···,$p_{10}$ = $v_{10}$}. In order to store a generalized neutrosophic soft set in a computer, we could represent it in the form of a table as shown below ( corresponding to the generalized neutrosophic soft set in the above example ). In this table, the entries $c_{ij}$ corresponding to the blouse $b_i$ and the parameter $e_j$, where $c_{ij}$ = (true-membership value of $b_i$, indeterminacy-membership value of $b_i$, falsity membership value of $b_i$) in F($e_j$). The table 1 represent the generalized neutrosophic soft set ( F, A ) described above.

Table 1: Tabular form of the GNSS ( F, A ).

| U | bright | cheap | costly | colorful |
|---|--------|-------|--------|----------|
| $b_1$ | ( 0.5, 0.6, 0.3 ) | ( 0.6, 0.3, 0.5 ) | ( 0.7, 0.4, 0.3 ) | ( 0.8, 0.1, 0.4 ) |
| $b_2$ | ( 0.4, 0.7, 0.2 ) | ( 0.7, 0.4, 0.3 ) | ( 0.6, 0.1, 0.2 ) | ( 0.4, 0.2, 0.6 ) |
| $b_3$ | ( 0.6, 0.2, 0.3 ) | ( 0.8, 0.1, 0.2 ) | ( 0.7, 0.2, 0.5 ) | ( 0.3, 0.6, 0.4 ) |
| $b_4$ | ( 0.7, 0.3, 0.2 ) | ( 0.7, 0.1, 0.3 ) | ( 0.5, 0.2, 0.6 ) | ( 0.4, 0.8, 0.5 ) |
| $b_5$ | ( 0.8, 0.2, 0.3 ) | ( 0.8, 0.3, 0.4 ) | ( 0.7, 0.3, 0.2 ) | ( 0.3, 0.5, 0.7 ) |

## Remark 3.4.

A generalized neutrosophic soft set is not a generalized neutrosophic set but a parametrized family of a generalized neutrosophic subsets.

## Definition 3.5

For two generalized neutrosophic soft sets ( F, A ) and ( G, B ) over the common universe U. We say that ( F, A ) is a generalized neutrosophic soft subset of ( G, B ) iff:

(i) A ⊂ B.
(ii) F(e) is a generalized neutrosophic subset of G(e).
    Or $T_{F(e)}(x) \leq T_{G(e)}(x)$, $I_{F(e)}(x) \leq I_{G(e)}(x)$, $F_{F(e)}(x) \geq F_{G(e)}(x)$, ∀ e ∈ A, x ∈ U.
We denote this relationship by ( F, A ) ⊆ ( G, B ).

( F, A ) is said to be generalized neutrosophic soft super set of ( G, B ) if ( G, B ) is a generalized neutrosophic soft subset of ( F, A ). We denote it by ( F, A ) ⊇ ( G, B ).

## Example 3.6

Let (F, A) and (G, B)  be two GNSS over the same universe U = {$o_1$, $o_2$, $o_3$, $o_4$, $o_5$}. The GNSS (F, A) describes the sizes of the objects whereas the GNSS ( G, B ) describes its surface textures. Consider the tabular representation of the GNSS ( F, A ) is as follows.

Table 2. Tabular form of the GNSS ( F,  A ).

| U | small | large | colorful |
|---|-------|-------|----------|
| $O_1$ | ( 0.4, 0.4, 0.6 ) | ( 0.3, 0.2, 0.7 ) | ( 0.4, 0.7, 0.5 ) |
| $O_2$ | ( 0.3, 0.5, 0.4 ) | ( 0.4, 0.7, 0.8 ) | ( 0.6, 0.3, 0.4 ) |
| $O_3$ | ( 0.6, 0.3, 0.5 ) | ( 0.3, 0.2, 0.6 ) | ( 0.4, 0.4, 0.8 ) |
| $O_4$ | ( 0.5, 0.1, 0.6 ) | ( 0.1, 0.6, 0.7 ) | ( 0.3, 0.5, 0.8 ) |
| $O_5$ | ( 0.3, 0.4, 0.4 ) | ( 0.3, 0.1, 0.6 ) | ( 0.5, 0.4, 0.4 ) |

The tabular representation of the GNSS ( G, B ) is given by table 3

Table 3. Tabular form of the GNSS ( G, B ).

| U | Small | Large | Colorful | Very smooth |
|---|-------|-------|----------|-------------|
| O1 | (0.6,  0.3, 0.3 ) | ( 0.5, 0.1, 0.4 ) | ( 0.5, 0.1, 0.4 ) | ( 0.1, 0.5, 0.4 ) |
| O2 | ( 0.7, 0.1, 0.2 ) | ( 0.4, 0.2, 0.3 ) | ( 0.7, 0.3, 0.2 ) | ( 0.5, 0.2, 0.3 ) |
| O3 | ( 0.6, 0.2, 0.5 ) | ( 0.7, 0.1, 0.4 ) | ( 0.6, 0.3, 0.3 ) | ( 0.2, 0.5, 0.4 ) |
| O4 | ( 0.8, 0.1, 0.4 ) | ( 0.3, 0.5, 0.4 ) | ( 0.4, 0.3, 0.7 ) | ( 0.4, 0.4, 0.5 ) |
| O5 | ( 0.5, 0.2, 0.2 ) | ( 0.4, 0.1, 0.5 ) | ( 0.6, 0.2, 0.3 ) | ( 0.5, 0.6, 0.3 ) |

Clearly, by definition 3.5 we have ( F, A )　　( G, B ).

## Definition 3.7

Two GNSS ( F, A ) and ( G, B ) over the common universe U are said to be generalized neutrosophic soft equal if ( F, A ) is generalized neutrosophic soft subset of ( G, B ) and ( G, B ) is generalized neutrosophic soft subset of ( F, A ) which can be written as ( F, A )= ( G, B ).

## Definition 3.8

Let E = {$e_1,e_2,\cdots,e_n$} be a set of parameters. The NOT set of E is denoted by ⌉E is defined by ⌉E = {$_\rfloor e_1, _\rfloor e_2, \cdots, _\rfloor e_n$}, where $_\rfloor e_i$ = not $e_i$,　i ( it may be noted that ⌉ and $_\rfloor$ are different operators ).

## Example 3.9

Consider the example 3.3. Here ⌉E = { not bright, not cheap, not costly, not colorful }.

## Definition 3.10

The complement of  generalized neutrosophic soft set ( F, A ) is denoted by $(F,A)^c$ and is defined by $(F,A)^c$= ($F^c$, ⌉A), where $F^c$ :⌉A → N(U) is a mapping given by $F^c$( $ɑ$ ) = generalized neutrosophic soft complement with $T_{F^c(x)} = F_{F(x)}, I_{F^c(x)} = I_{F(x)}$ and $F_{F^c(x)} = T_{F(x)}$.

## Example 3.11

As an illustration consider the example presented in the example 3.2. the complement $(F,A)^c$ describes the 'not attractiveness of the blouses'. Is given below.

F( not bright) = {< $b_1$, 0.3, 0.6, 0.5 >,< $b_2$, 0.2, 0.7, 0.4 >,< $b_3$, 0.3, 0.2, 0.6 >,
                 < $b_4$, 0.2, 0.3, 0.7 >< $b_5$, 0.3, 0.2, 0.8 >}.
F( not cheap ) = {< $b_1$, 0.5, 0.3, 0.6 >,< $b_2$, 0.3, 0.4, 0.7 >,< $b_3$, 0.2, 0.1, 0.8 >,
                 < $b_4$, 0.3, 0.1, 0.7 >,< $b_5$, 0.4, 0.3, 0.8 >}.
F( not costly ) = {< $b_1$, 0.3, 0.4, 0.7 >,< $b_2$, 0.2, 0.1, 0.6 >,< $b_3$, 0.5, 0.2, 0.7 >,
                 < $b_4$, 0.6, 0.2, 0.5 >, < $b_5$, 0.2, 0.3, 0.7 >}.
F( not colorful ) = {< $b_1$, 0.4, 0.1, 0.8 >, < $b_2$, 0.6, 0.2, 0.4 >,< $b_3$, 0.4, 0.6, 0.3 >,
                 < $b_4$, 0.5, 0.8, 0.4 >< $b_5$, 0.7, 0.5, 0.3 >}.

## Definition 3.12

The generalized neutrosophic soft set (F,A) over U is said to be empty or null generalized neutrosophic soft (with respect to the set of parameters) denoted by $_A$ or ( ,A) if $T_{F(e)}(m)$ = 0,$F_{F(e)}(m)$ = 0 and $I_{F(e)}(m)$ = 0,　m　U,　e　A.

## Example 3.13

Let U = {$b_1,b_2,b_3,b_4,b_5$}, the set of five blouses be considered as the universal set and A = { Bright, Cheap, Colorful } be the set of parameters that characterizes the blouses. Consider the generalized neutrosophic soft set ( F, A ) which describes the cost of the blouses and

F(bright)={< $b_1$, 0, 0, 0 >,< $b_2$,0, 0, 0 >,< $b_3$, 0, 0, 0 >,< $b_4$, 0 ,0, 0 >, < $b_5$, 0, 0, 0 >},

F(cheap)={< $b_1$, 0, 0, 0 >,< $b_2$,0, 0, 0 >,< $b_3$, 0, 0, 0 >,< $b_4$, 0, 0, 0 >, < $b_5$, 0, 0, 0 >},
F(colorful)={< $b_1$, 0, 0, 0 >,< $b_2$, 0, 0 ,0 >,< $b_3$, 0, 0, 0 >, < $b_4$, 0, 0, 0 >,< $b_5$, 0, 0, 0 >}.
Here the NGNSS ( F, A ) is the null generalized neutrosophic soft set.

**Definition 3.14**. **Union of Two Generalized Neutrosophic Soft Sets.**

Let (F, A) and (G, B) be two GNSS over the same universe U. Then the union of (F, A) and (G, B) is denoted by '(F, A)    (G, B)' and is defined by (F, A)    (G, B) = (K, C), where C = A   B and the truth-membership, indeterminacy-membership and falsity-membership of ( K, C) are as follows:

$T_{K(e)}(m)$   = $T_{F(e)}(m)$, if e    A − B
           = $T_{G(e)}(m)$, if e    B − A
           = max ($T_{F(e)}(m)$, $T_{G(e)}(m)$), if e    A    B.
$I_{K(e)}(m)$ = $I_{F(e)}(m)$, if e    A − B
           = $I_{G(e)}(m)$, if e    B − A
           = min ($I_{F(e)}(m)$, $I_{G(e)}(m)$)), if e    A    B.
$F_{K(e)}(m)$   = $F_{F(e)}(m)$, if e    A − B
           = $F_{G(e)}(m)$, if e    B − A
           = min ($F_{F(e)}(m)$, $F_{G(e)}(m)$), if e    A    B.

**Example 3.15**. Let ( F, A ) and ( G, B ) be two GNSS over the common universe U. Consider the tabular representation of the GNSS ( F, A ) is as follow:

Table 4.Tabular form of the GNSS ( F, A ).

|  | Bright | Cheap | Colorful |
|---|---|---|---|
| $b_1$ | ( 0.6, 0.3, 0.5 ) | ( 0.7, 0.3, 0.4 ) | ( 0.4, 0.2, 0.6 ) |
| $b_2$ | ( 0.5, 0.1, 0.8 ) | ( 0.6, 0.1, 0.3 ) | ( 0.6, 0.4, 0.4 ) |
| $b_3$ | ( 0.7, 0.4, 0.3 ) | ( 0.8, 0.3, 0.5 ) | ( 0.5, 0.7, 0.2 ) |
| $b_4$ | ( 0.8, 0.4, 0.1 ) | ( 0.6, 0.3, 0.2 ) | ( 0.8, 0.2, 0.3 |
| $b_5$ | ( 0.6, 0.3, 0.2 ) | ( 0.7, 0.3, 0.5 ) | ( 0.3, 0.6, 0.5 |

The tabular representation of the GNSS ( G, B ) is as follow:

Table 5. Tabular form of the GNSS ( G, B ).

| U | Costly | Colorful |
|---|---|---|
| $b_1$ | ( 0.6, 0.2, 0.3) | ( 0.4, 0.6, 0.2 ) |
| $b_2$ | ( 0.2, 0.7, 0.2 ) | ( 0.2, 0.8, 0.3 ) |
| $b_3$ | ( 0.3, 0.6, 0.5 ) | ( 0.6, 0.3, 0.4 ) |
| $b_4$ | ( 0.8, 0.4, 0.1 ) | ( 0.2, 0.8, 0.3 ) |
| $b_5$ | ( 0.7, 0.1, 0.4 ) | ( 0.5, 0.6, 0.4 ) |

Using definition 3.12 the union of two  GNSS (F, A ) and ( G, B ) is ( K, C ) can be represented as follow.

Table 6. Tabular form of the GNSS ( K, C ).

| U | Bright | Cheap | Colorful | Costly |
|---|--------|-------|----------|--------|
| $b_1$ | ( 0.6, 0.3, 0.5 ) | ( 0.7, 0.3, 0.4 ) | ( 0.4, 0.2, 0.2 ) | ( 0.6, 0.2, 0.3 ) |
| $b_2$ | ( 0.5, 0.1, 0.8 ) | ( 0.6, 0.1, 0.3 ) | ( 0.6, 0.4, 0.3 ) | ( 0.2, 0.7, 0.2 ) |
| $b_3$ | ( 0.7, 0.4, 0.3 ) | ( 0.8, 0.3, 0.5 ) | ( 0.6, 0.3, 0.2 ) | ( 0.3, 0.6, 0.5 ) |
| $b_4$ | ( 0.8, 0.4, 0.1 ) | ( 0.6, 0.3, 0.2 ) | ( 0.8, 0.2, 0.3 ) | ( 0.8, 0.4, 0.1 ) |
| $b_5$ | ( 0.6, 0.3, 0.2 ) | ( 0.7, 0.3, 0.5 ) | ( 0.5, 0.6, 0.4 ) | ( 0.7, 0.1, 0.4 ) |

**Definition 3.16. Intersection of Two Generalized Neutrosophic Soft Sets.**

Let (F,A) and (G,B) be two GNSS over the same universe U such that A ∩ B ≠ 0. Then the intersection of (F, A) and ( G, B) is denoted by '( F, A) ∩ (G, B)' and is defined by ( F, A) ∩ ( G, B ) = ( K, C), where C =A ∩ B and the truth-membership, indeterminacy membership and falsity-membership of ( K, C ) are related to those of (F, A) and (G, B) by:

$T_{K(e)}(m) = \min (T_{F(e)}(m), T_{G(e)}(m))$
$I_{K(e)}(m) = \min (I_{F(e)}(m), I_{G(e)}(m))$
$F_{K(e)}(m) = \max (F_{F(e)}(m), F_{G(e)}(m))$, for all e ∈ C.

**Example 3.17**. Consider the above example 3.15. The intersection of ( F, A ) and ( G, B ) can be represented into the following table :

Table 7. Tabular form of the GNSS ( K, C ).

| U | Colorful |
|---|----------|
| $b_1$ | ( 0.4, 0.2, 0.6 ) |
| $b_2$ | ( 0.2, 0.4, 0.4 ) |
| $b_3$ | ( 0.6, 0.3, 0.4 ) |
| $b_4$ | ( 0.8, 0.2, 0.3 ) |
| $b_5$ | ( 0.3, 0.6, 0.5 ) |

**Proposition 3.18.** If (F, A) and (G, B) are two GNSS over U and on the basis of the operations defined above , then:

(1)   (F, A) ∪ (F, A) = (F, A).
       (F, A) ∩ (F, A) = (F, A).
(2)   (F, A) ∪ (G, B) = (G, B) ∪ (F, A).
       (F, A) ∩ (G, B) = (G, B) ∩ (F, A).
(3)   (F, A) ∪ Φ = (F, A).
(4)   (F, A) ∩ Φ = Φ.
(5)   [(F, A)$^c$]$^c$ = (F, A).
Proof. The proof of the propositions 1 to 5 are obvious.

**Proposition 3.19.** If ( F, A ), ( G, B ) and ( K, C ) are three GNSS over U, then:

(1)     (F, A) ∪ [(G, B) ∩ (K, C)] = [(F, A) ∪ (G, B)] ∩ (K, C).

(2)      (F, A)    [(G, B)    (K, C)] = [(F, A)    (G, B)]    (K, C).

(3)      (F, A)    [(G, B)    (K, C)] = [(F, A)    (G, B)]    [(F, A)    (K, C)].

(4)      (F, A)    [(G, B)    (K, C)] = [(H, A)    (G, B)]    [(F, A)    (K, C)].

**Example 3.20.** Let (F,A) ={ $b_1$ ,0.6, 0.3, 0. 1    , $b_2$, 0.4, 0.7, 0. 5) ,($b_3$, 0.4, 0.1, 0.8) } , (G,B) =|($b_1$, 0.2, 0.2, 0.6), ($b_2$ ,0.7, 0.2, 0.4), ($b_3$,0.1, 0.6, 0.7) } and (K,C) =|($b_1$, 0.3, 0.8, 0.2), $b_2$,  0.4, 0.1, 0.6) , $b_3$,0.9, 0.1, 0.2)} be three GNSS of U, Then:

(F, A)    (G, B) = {  $b_1$, 0.6, 0.2, 0.1    , $b_2$, 0.7, 0.2 ,0.4    , $b_3$,0.4, 0.1, 0.7   }.

(F, A)    (K, C) = {  $b_1$, 0.6, 0.3, 0.1   , $b_2$, 0.4, 0.1, 0.5    , $b_3$,0.9, 0.1, 0.2   }.

(G, B)    (K, C)] = {  $b_1$, 0.2, 0.2, 0.6   , $b_2$,0.4, 0.1, 0.6  , $b_3$, 0.1, 0.1, 0.7   }.

 (F, A)    [(G, B)    (K, C)] = {  $b_1$, 0.6, 0.2, 0.1   , $b_2$,0.4, 0.1, 0.5    , $b_3$,0.4, 0.1, 0.7   }.

 [(F, A)    (G, B)]    [(F, A)    (K, C)] = {  $b_1$,0.6,  0.2,  0.1 , $b_2$,  0.4,  0.1,  0.5 , $b_3$,  0.4, 0.1,0.7  }.

Hence distributive (3)  proposition verified.

Proof, can be easily proved from definition 3.14.and 3.16.

**Definition 3.21**. **AND Operation on Two Generalized Neutrosophic Soft Sets**.

Let ( F, A ) and ( G, B ) be two GNSS over the same universe U. then ( F, A ) ''AND ( G, B) denoted by '( F, A )    ( G, B ) and is defined by ( F, A )    ( G, B ) = ( K, A × B ), where K( α , β )=F( α )    B( β ) and the truth-membership, indeterminacy-membership and falsity-membership of ( K, A×B ) are as follows:

$T_{K(\ ,\ )}(m) = \min(T_{F(\ )}(m), T_{G(\ )}(m)), I_{K(\ ,\ )}(m) = \min(I_{F(\ )}(m), I_{G(\ )}(m))$

$F_{K(\ ,\ )}(m) = \max(F_{F(\ )}(m), F_{G(\ )}(m)),$   α ∈ A,   β ∈ B.

**Example 3.22**. Consider the same example 3.15 above. Then the tabular representation of (F,A) and( G, B ) is as follow:

Table 8: Tabular representation of the GNSS ( K, A × B).

| u | (bright, costly) | (bright, Colorful) | (cheap, costly) |
|---|---|---|---|
| $b_1$ | ( 0.6, 0.2, 0.5 ) | ( 0.4, 0.3, 0.5 ) | ( 0.6, 0.2, 0.4 ) |
| $b_2$ | ( 0.2, 0.1, 0.8 ) | ( 0.2, 0.1, 0.8 ) | ( 0.2, 0.1, 0.3 ) |
| $b_3$ | ( 0.3, 0.4, 0.5 ) | ( 0.6, 0.3, 0.4 ) | ( 0.3, 0.3, 0.5 ) |
| $b_4$ | ( 0.8, 0.4, 0.1 ) | ( 0.2, 0.4, 0.3 ) | ( 0.6, 0.3, 0.2 ) |
| $b_5$ | ( 0.6, 0.1, 0.4 ) | ( 0.5, 0.3, 0.4 ) | ( 0.7, 0.1, 0.5) |

| u | (cheap, colorful) | (colorful, costly) | (colorful, colorful) |
|---|---|---|---|
| $b_1$ | ( 0.4, 0.3, 0.4 ) | ( 0.4, 0.2, 0.6 ) | ( 0.4, 0.2, 0.6 ) |
| $b_2$ | ( 0.2, 0.1, 0.3 ) | ( 0.2, 0.4, 0.4 ) | ( 0.2, 0.4, 0.4 ) |
| $b_3$ | ( 0.6, 0.3, 0.5 ) | ( 0.3, 0.6, 0.5 ) | ( 0.5, 0.3, 0.4 ) |
| $b_4$ | ( 0.2, 0.3, 0.3 ) | ( 0.8,0.2, 0.3 ) | ( 0.2, 0.2, 0.3 ) |
| $b_5$ | ( 0.5, 0.3, 0.5 ) | ( 0.3, 0.1, 0.5 ) | ( 0.3, 0.6, 0.5 ) |

**Definition 3.23**. If (F, A) and (G, B) be two GNSS over the common universe U then '(F, A) OR (G, B)' denoted by (F, A) ∪ (G, B) is defined by ( F, A) ∪ (G, B ) = (O, A×B), where, the truth-membership, indeterminacy membership and falsity-membership of O( , ) are given as follows:

$$T_{O(\alpha,\beta)}(m) = {}^{max}(T_{F(\alpha)}(m), T_{G(\beta)}(m))$$

$$I_{O(\alpha,\beta)}(m) = {}^{min}(I_{F(\alpha)}(m), I_{G(\beta)}(m))$$

$$F_{O(\alpha,\beta)}(m) = \min(F_{F(\alpha)}(m), F_{G(\beta)}(m)), \forall \alpha \in A, \forall \beta \in B.$$

**Example 3.24.** Consider the same example 3.14 above. Then the tabular representation of ( F, A ) OR ( G, B ) is as follow:

Table 9: Tabular representation of the GNSS ( O, A × B).

| u | (bright, costly) | (bright, colorful) | (cheap, costly) |
|---|---|---|---|
| $b_1$ | ( 0.6, 0.2, 0.3 ) | ( 0.6, 0.3, 0.2 ) | ( 0.7, 0.2, 0.3 ) |
| $b_2$ | ( 0.5, 0.1, 0.2 ) | ( 0.5, 0.1, 0.3 ) | ( 0.6, 0.1, 0.2 ) |
| $b_3$ | ( 0.7, 0.4, 0.3 ) | ( 0.7, 0.3, 0.3 ) | ( 0.8 ,0.3, 0.5 ) |
| $b_4$ | ( 0.8, 0.4, 0.1 ) | ( 0.8, 0.4, 0.1 ) | ( 0.8, 0.3, 0.1 ) |
| $b_5$ | ( 0.7, 0.1, 0.2 ) | ( 0.6, 0.3, 0.4 ) | ( 0.7, 0.1, 0.4 ) |

| u | (cheap, colorful) | (colorful, costly) | (colorful, colorful) |
|---|---|---|---|
| $b_1$ | ( 0.7, 0.3, 0.2 ) | ( 0.6, 0.2, 0.3 ) | ( 0.4, 0.2, 0.2 ) |
| $b_2$ | ( 0.6, 0.1, 0.3 ) | ( 0.6, 0.4, 0.2 ) | ( 0.6, 0.4, 0.3 ) |
| $b_3$ | ( 0.8, 0.3, 0.4 ) | ( 0.5, 0.6, 0.2 ) | ( 0.5, 0.7, 0.2 ) |
| $b_4$ | ( 0.6, 0.3, 0.2 ) | ( 0.8, 0.2, 0.1 ) | ( 0.8, 0.2, 0.3 ) |
| $b_5$ | ( 0.7, 0.3, 0.4 ) | ( 0.7, 0.1, 0.4 ) | ( 0.5, 0.6, 0.4) |

**Proposition 3.25.** If ( F, A ) and ( G, B ) are two GNSS over U, then **:**
(1) [(F, A) ∪ (G, B)]$^c$ = (F,A)$^c$ ∩ (G, B)$^c$
(2) [(F, A) ∩ (G, B)]$^c$ = (F,A)$^c$ ∪ (G, B)$^c$

**Proof 1.**

Let (F, A)={<b, $T_{F(x)}(b)$, $I_{F(x)}(b)$, $F_{F(x)}(b)$>|b ∈ U }
and (G, B) = {< b, $T_{G(x)}(b)$, $I_{G(x)}(b)$, $F_{G(x)}(b)$ > |b ∈ U }
be two GNSS over the common universe U. Also let (K, A × B) = (F, A) ∪ (G, B),
where, K( , ) = F( ) ∩ G( ) for all ( , ) ∈ A × B then
K( , ) = {< b, min($T_{F( )}(b)$,$T_{G( )}(b)$), min($I_{F( )}(b)$,$I_{G( )}(b)$), max($F_{F( )}(b)$,$F_{G( )}(b)$) >| b ∈ U }.

Therefore,
[(F,A) ∪ (G, B)]$^c$ = (K, A × B)$^c$
= {< b, max($F_{F( )}(b)$,$F_{G( )}(b)$), min($I_{F( )}(b)$,$I_{G( )}(b)$), min($T_{F( )}(b)$,$T_{G( )}(b)$) >|b ∈ U }.
Again

$(F, A)^c \quad (G, B)^c$

$= \{< b, \max(F_{F^c(\ )}(b)), F_{G^c(\ )}(b)), \min(I_{F^c(\ )}(b), I_{G^c(\ )}(b)), \min(T_{F^c(\ )}(b), T_{G^c(\ )}(b)) >| b \quad U\}$.

$= \{< b, \min(T_{F(\ )}(b), T_{G(\ )}(b)), \min(I_{F(\ )}(b), I_{G(\ )}(b)), \max(F_{F(\ )}(b), F_{G(\ )}(b)) >| b \quad U\}^c$.

$= \{< b, \max(F_{F(\ )}(b), F_{G(\ )}(b)), \min(I_{F(\ )}(b), I_{G(\ )}(b)), \min(T_{F(\ )}(b), T_{G(\ )}(b)) >| b \quad U\}$.

It follows that $[(F, A) \quad (G, B)]^c = (F, A)^c \quad (G, B)^c$.

**Proof 2.**

Let $( F, A ) = \{< b, T_{F(x)}(b), I_{F(x)}(b), F_{F(x)}(b) >|b \quad U\}$ and

$(G, B) = \{< b, T_{G(x)}(b), I_{G(x)}(b), F_{G(x)}(b) > |b \quad U\}$ be two GNSS over the common universe U. Also let $(O, A \times B) = (F, A) \quad (G, B)$, where, $O (\ , \ ) = F(\ ) \quad G(\ )$ for all $(\ , \ ) \quad A \times B$. Then

$O(\ , \ ) = \{< b, \max(T_{F(\ )}(b), T_{G(\ )}(b)), \min(I_{F(\ )}(b), I_{G(\ )}(b)), \min(F_{F(\ )}(b), F_{G(\ )}(b)) > |b \quad U\}$.

$[(F, A) \quad (G, B)]^c = (O, A \times B)^c = \{< b, \min(F_{F(\ )}(b), F_{G(\ )}(b)), \min(I_{F(\ )}(b), I_{G(\ )}(b)), \max(T_{F(\ )}(b), T_{G(\ )}(b)) > |b \quad U\}$.

Again

$(H, A)^c \quad (G, B)^c$

$= \{< b, \min(F_{F^c(\ )}(b), F_{G^c(\ )}(b)), \min(I_{F^c(\ )}(b), I_{G^c(\ )}(b)), \max(T_{F^c(\ )}(b), T_{G^c(\ )}(b)), >| b \quad U\}$.

$= \{< b, \max(T_{F(\ )}(b), T_{G(\ )}(b)), \min(I_{F^c(\ )}(b), I_{G^c(\ )}(b)), \min(F_{F(\ )}(b), F_{G(\ )}(b)) >| b \quad U\}^c$.

$= \{< b, \min(F_{F(\ )}(b), F_{G(\ )}(b)), \min(I_{F(\ )}(b), I_{G(\ )}(b)), \max(T_{F(\ )}(b), T_{G(\ )}(b)) >| b \quad U\}$.

It follows that $[(F, A) \quad (G, B)]^c = (F, A)^c \quad (G, B)^c$.

# 4. An Application of Generalized Neutrosophic Soft Set in a Decision Making Problem

To see an application of the concept of generalized neutrosophic soft set:

Let us consider the generalized neutrosophic soft set $S = (F,P)$ (see also Table 10 for its tabular representation), which describes the "attractiveness of the blouses" that Mrs. X is going to buy. on the basis of her m number of parameters $(e_1, e_2, \ldots, e_m)$ out of n number of blouses $(b_1, b_2, \ldots, b_n)$. We also assume that corresponding to the parameter $e_j$ $(j = 1, 2, \cdots, m)$ the performance value of the blouse $b_i$ $(i = 1, 2, \cdots, n)$ is a tuple $p_{ij} = (T_{F(ej)} (b_i), I_{F(ej)} (b_i), T_{F(ej)} (b_i))$, such that for a fixed i that values $p_{ij}$ $(j = 1, 2, \cdots, m)$ represents a generalized neutrosophic soft set of all the n objects. Thus the performance values could be arranged in the form of a matrix called the 'criteria matrix'. The more are the criteria values, the more preferability of the corresponding object is. Our problem is to select the most suitable object i.e. The object which dominates each of the objects of the spectrum of the parameters $e_j$. Since the data are not crisp but generalized neutrosophic soft the selection is not straightforward. Our aim is to find out the most suitable blouse with the choice parameters for Mrs. X. The blouse which is suitable for Mrs. X need not be suitable for Mrs. Y or Mrs. Z, as the selection is dependent on the choice parameters of each buyer. We use the technique to calculate the score for the objects.

## 4.1. Definition: Comparison matrix.

The Comparison matrix is a matrix whose rows are labeled by the object names of the universe such as $b_1, b_2, \cdots, b_n$ and the columns are labeled by the parameters $e_1, e_2, \cdots, e_m$. The entries are $c_{ij}$, where $c_{ij}$ is the number of parameters for which the value of $b_i$ exceeds or is equal to the value $b_j$. The entries are calculated by $c_{ij} = a + d - c$, where 'a' is the integer calculated as 'how many times $T_{bi}$ ($e_j$) exceeds or equal to $T_{bk}$ ($e_j$)', for $b_i$    $b_k$,    $b_k$ ∈ U, 'd' is the integer calculated as 'how many times $I_{bi(ej)}$ exceeds or equal to $I_{bk(ej)}$', for $b_i$   $b_k$,    $b_k$ ∈ U and 'c' is the integer 'how many times $F_{bi(ej)}$ exceeds  or equal to $F_bk(ej)$', for $b_i$   $b_k$,   $b_k$ ∈ U.

**Definition 4.2.**  Score of an object.  The score of an object $b_i$  is $S_i$  and is calculated as :

$$S_i = \sum_j c_{ij}$$

Now the  algorithm for most appropriate selection of an object will be as follows.

Algorithm

(1) input the  generalized neutrosophic Soft Set ( F, A).

(2) input P, the choice parameters of Mrs. X which is a subset of A.

(3) consider the GNSS ( F, P) and write it in tabular form.

(4) compute the comparison matrix of the GNSS ( F, P).

(5) compute the score $S_i$ of $b_i$,   i.

(6) find $S_k$ = maxi $S_i$

(7) if k has more than one value then any one of $b_i$  may be chosen.

To illustrate the basic idea of the algorithm, now we apply it to the generalized neutrosophic soft set based decision making problem.

Suppose the wishing parameters for Mrs. X where P={ Cheap, Colorful, Polystyrening, costly , Bright }.

Consider the GNSS ( F, P )  presented into the following table.

Table 10. Tabular form of the GNSS (F, P).

| U | Cheap | Colorful | Polystyreneing | costly | Bright |
|---|---|---|---|---|---|
| $b_1$ | ( 0.6, 0.3,  0.4 ) | ( 0.5, 0.2, 0.6 ) | ( 0.5, 0.3, 0.4 ) | ( 0.8, 0.2, 0.3 ) | ( 0.6,0.3, 0.2 ) |
| $b_2$ | ( 0.7, 0.2,  0.5 ) | ( 0.6, 0.3, 0.4 ) | ( 0.4, 0.2, 0.6 ) | ( 0.4, 0.8, 0.3 ) | ( 0.8,0.1, 0.2 ) |
| $b_3$ | ( 0.8, 0.3, 0.4 ) | ( 0.8, 0.5, 0.1 ) | ( 0.3, 0.5, 0.6 ) | ( 0.7, 0.2, 0.1 ) | ( 0.7,0.2, 0.5 ) |
| $b_4$ | ( 0.7, 0.5,  0.2 ) | ( 0.4, 0.8, 0.3 ) | ( 0.8, 0.2, 0.4 ) | ( 0.8, 0.3, 0.4 ) | ( 0.8,0.3, 0.4 ) |
| $b_5$ | ( 0.3, 0.8, 0.4 ) | ( 0.3, 0.6, 0.1 ) | ( 0.7, 0.3, 0.2 ) | ( 0.6,0.2, 0.4 ) | ( 0.6,0.4, 0.2 ) |

The comparison-matrix of the above GNSS ( F, P) is represented as follow:

Table 11. Comparison matrix of the GNSS ( F, P ).

| U | Cheap | Colorful | Polystyreneing | costly | Bright |
|---|---|---|---|---|---|
| $b_1$ | 0 | -2 | 3 | 0 | 2 |
| $b_2$ | -1 | 1 | -2 | 2 | 2 |
| $b_3$ | 3 | 5 | 0 | 4 | -1 |
| $b_4$ | 6 | 3 | 3 | 3 | 4 |
| $b_5$ | 7 | 2 | 6 | -1 | 3 |

Next we compute the score for each $b_i$ as shown below:

| U | Score ($S_i$) |
|---|---|
| $b_1$ | 3 |
| $b_2$ | 2 |
| $b_3$ | 11 |
| $b_4$ | 19 |
| $b_5$ | 17 |

Clearly, the maximum score is the score 19, shown in the table above for the blouse $b_4$.
Hence the best decision for Mrs. X is to select $b_4$, followed by $b_5$ .

## 5. CONCLUSIONS

In this article ,our main intention was to incorporate the generalized neutrosophic set proposed by A. A. Salama [7] in soft sets introduced by Molodtsov [4] considering the fact that the parameters ( which are words or sentences ) are mostly generalized neutrosophic set; but both the concepts deal with imprecision, We have also defined some operations on GNSS and present an application of GNSS in a decision making problem. Finally, we hope that our model opened a new direction, new path of thinking to engineers, mathematicians, computer scientist and many other in various tests.

## ACKNOWLEDGEMENTS

Our special thanks to the anonymous referee and the editor of this journal of their valuable comments and suggestions which have improved this paper.

## Authors

**Broumi Said** is an administrator of university of Hassan II-Mohammedia - Casablanca. He worked in University for five years. He received his Master in industrial automatic from the University of Hassan II –Ain chok. His research concentrates on soft set theory, fuzzy theory, intuitionistic fuzzy theory , neutrosophic theory, and control of systems. 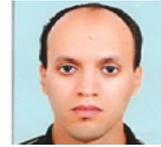